\documentclass[runningheads]{llncs}

\usepackage[T1]{fontenc}
\usepackage[utf8]{inputenc}
\usepackage{amsmath,amsfonts}
\usepackage{algorithm}
\usepackage{algorithmic}
\usepackage{array}
\usepackage{graphicx}
\usepackage{booktabs}
\usepackage{makecell}
\usepackage{textcomp}
\usepackage{xspace}
\usepackage{url}
\usepackage{hyperref}
\usepackage[section]{placeins}
\begin{document}

\title{A Novel Machine Learning Approach for Central Nervous System Tumor Classification from DNA Methylation}
\titlerunning{Machine Learning for CNS Tumor Classification}
\author{Paulo R. Ferreira-Jr.\inst{1}
\and Lucas Coutinho Freitas\inst{1}
\and Laís dos Santos Gonçalves\inst{1}
\and William Borges Domingues\inst{1}
\and Lucas Petitemberte de Souza\inst{1}
\and Mariana B. Michalowski\inst{2}
\and Vinicius F. Campos\inst{1}}
\authorrunning{Ferreira-Jr., P. R. et al.}
\institute{Universidade Federal de Pelotas, Pelotas, Rio Grande do Sul, Brazil\\
\and Hospital de Clínicas de Porto Alegre, Porto Alegre, Rio Grande do Sul, Brazil\\
\email{paulo.ferreira@inf.ufpel.edu.br}}

\maketitle

\begin{abstract}
DNA methylation profiling has become a powerful approach for central nervous system (CNS) tumor classification, yet important challenges remain regarding cross-cohort transferability, methodological correctness, and robust multiclass evaluation. In this work, we propose a novel and methodologically rigorous machine-learning approach for methylation-based CNS tumor classification that combines Sparse Random Projection for dimensionality reduction with multinomial logistic regression for classification. We evaluate the proposed approach in the same general experimental setting established by a widely used reference classifier. On the 2,801-sample reference cohort, our method achieves a mean accuracy of 96\% under stratified 3-fold cross-validation. On the independent 1,104-sample clinical evaluation cohort, it reaches 86\% accuracy at the 91-class level and 93\% when predictions are evaluated at the methylation class family level. These results improve upon the corresponding state-of-the-art reference figures of 82\% class-level concordance and 88\% family-level concordance, yielding absolute gains of approximately 4 and 5 percentage points, respectively. This improvement is clinically relevant: in a diagnostic setting, a 5-point increase in correct tumor classification can directly affect cancer subtype assignment and, in turn, influence treatment selection and downstream clinical decision-making. Our results show that the proposed model, grounded in stronger methodological practice in machine learning, consistently outperforms the previous state of the art across evaluation settings and can materially improve the reliability of CNS tumor classification.

\end{abstract}

\section{Introduction}

Accurate classification of central nervous system (CNS) tumors is vital for guiding diagnosis, prognosis, and personalized treatment decisions. DNA methylation profiling has emerged as a powerful tool for improving tumor classification beyond histopathology, enabling molecular-level precision in CNS tumor diagnosis. Among CNS tumors, medulloblastoma is a key example in which DNA methylation profiling has revolutionized classification and risk stratification. 

Capper \textit{et al.} \cite{Capper2018} developed a DNA methylation–based classifier that enabled the molecular classification of CNS tumors with remarkable accuracy. This classifier relies on genome-wide CpG methylation signatures to distinguish between 91 tumor methylation classes. Although this classifier is a research-use tool rather than a clinically validated test, methylation-based classification has since demonstrated straightforward diagnostic and clinical utility. It has been incorporated into routine neuropathology workflows in multiple centers.

Their training dataset comprises 2,801 reference samples, assayed on the Illumina 450K array. The workflow ranks 428,799 CpG probes using permutation-based random-forest importance across 43 probe sets (each comprising 100 trees) and then selects the 10,000 highest-ranked probes. Relying on a fixed probe subset may underrepresent biologically informative CpG sites that fall outside the selected set and could otherwise contribute additional predictive value. Notably, the manuscript does not provide a clear rationale for selecting 10,000 features specifically, nor does it report a sensitivity analysis assessing how performance varies with the number of selected probes.

The authors employ a random forest classifier with fixed hyperparameters (10,000 trees and downsampling to 8 samples per class). However, they do not provide detailed justification or a sensitivity analysis for these design choices. These parameters are central to the behavior and performance of ensemble models. Although they report calibrating the model's probabilities, standard calibration methods are designed for binary settings. They require adaptations for multiclass contexts, such as a ``match-vs-non-match'' formulation.

Regarding performance evaluation, the authors report an AUC of 0.99, indicating high predictive accuracy. Nevertheless, this metric was derived from a binarised ``match-vs-non-match'' formulation based on thresholds, rather than from the original 91-class prediction task. As a result, the reported AUC reflects the model's discrimination rather than its actual multiclass performance. Including multiclass evaluation metrics would provide a more comprehensive understanding of model behavior and enhance the interpretability and clinical robustness of the reported results.

The authors also conducted a clinical evaluation of 1,104 prospective cases. The classifier demonstrated progressive concordance with histopathological diagnosis, increasing from 76\% at the class level to 88\% when broader methylation class families (MCFs) were considered. This reflects an effort to capture biological relatedness among tumor types, though it also changes the granularity of the evaluation. At this point in the analysis, however, the evaluation framework was broadened from individual methylation classes to methylation class families (MCFs), which group related entities into broader diagnostic categories. 

Under this higher-level classification, predictions within the same MCF were considered concordant, resulting in a reported performance of 88\%. This approach reflects an effort to capture the biological and clinical proximity between tumor entities. However, it also introduces a shift in evaluation granularity that should be carefully contextualized when interpreting results.

Motivated by the methodological limitations of the available classifier discussed above, we propose a novel machine learning approach that follows best practices for high-dimensional data and enhances the correctness, reproducibility, and interpretability of the results.

We reduce the feature space using Sparse Random Projection (SRP), an efficient dimensionality-reduction technique suitable for datasets with hundreds of thousands of variables and relatively few samples per class. SRP preserves relevant data structure while improving model tractability. This step computes compact features from the probe-level measurements, thereby avoiding reliance on a fixed, hand-selected subset of CpG sites. 

For classification, we employ logistic regression, a well-established model for high-dimensional, low-sample-size problems, and we do not apply downsampling. 

We evaluate our proposed workflow using stratified cross-validation on the reference cohort and validate it on the independent clinical dataset, reporting comprehensive performance metrics and per-class analyses. In the same experimental setting described by \cite{Capper2018}, our approach achieves a superior performance.

The remainder of this paper is structured as follows. Section \ref{S-ReWork} presents the foundations of methylation-based tumor classification and machine learning best practices. Section \ref{S-Crititical} analyses the methodology proposed by \cite{Capper2018}, addressing issues in feature selection, calibration, and validation. Section \ref{S-Proposed} introduces our proposed workflow. Section \ref{S-ExpResult} presents the experimental setup and results, and Section \ref{S-conclusion} concludes the paper.

\section{Foundations on DNA Methylation}
\label{S-ReWork}

DNA methylation is a central mechanism of epigenetic regulation, classically described as heritable regulation that alters gene function without changing the underlying DNA sequence. In mammals, methylation predominantly occurs at the 5-carbon position of cytosine, forming 5-methylcytosine (5mC) at CpG dinucleotides. Another form of methylation can occur at the N6 position of adenine (6mA), producing N6-methyladenosine. It is estimated that approximately 60–80\% of CpG sites in the human genome are methylated. The functional impact of 5mC is highly context-dependent: similar methylation patterns can be associated with transcriptional activation or repression, depending on the genomic location and gene-specific regulatory context \cite{Guo2013NonCpGMethylationBrain}.

DNA methylation plays a crucial role in various biological processes, including embryonic development, genomic imprinting, X-chromosome inactivation, and the maintenance of chromosomal stability \cite{CapellBerger2013GenomeWideEpigenetics}. Epigenetic regulation is mediated through various mechanisms, including direct cytosine methylation (5mC), post-translational histone modifications (such as methylation, acetylation, phosphorylation, and ubiquitination), and regulation by non-coding RNAs \cite{GreerShi2012HistoneMethylation}. Many of these processes are reversible and can operate in a coordinated fashion.

In cancer epigenomics, identifying candidate epigenetic biomarkers typically requires detecting aberrantly methylated regions, including differential methylation patterns that appear hypomethylated (low methylation) or hypermethylated (high methylation) relative to other probes and samples. Methylation levels are commonly quantified using $beta$ values and $M$ values. $Beta$ values represent the proportion of methylated signal intensity and are bounded between $0$ and $1$; values at or below $0.2$ are often interpreted as hypomethylation, whereas values at or above $0.8$ are interpreted as hypermethylation. $M$ values, defined as the log-ratio of methylated to unmethylated signal intensities, are often preferred for statistical analyses due to their improved statistical properties. Values less than or equal to $-2$ typically indicate hypomethylation, and values greater than or equal to $2$ indicate hypermethylation \cite{ChenLinFann2016DMRMethods}.

DNA methylation microarrays quantify methylation at hundreds of thousands of CpG loci by combining bisulfite conversion with probe-based hybridization chemistry. In brief, extracted genomic DNA is bisulfite-treated so that unmethylated cytosines are converted to uracils (and read as thymines after amplification), whereas methylated cytosines remain unchanged. The converted DNA is then amplified, fragmented, and hybridized to array probes that interrogate each CpG using methylated and unmethylated channels (for Infinium assays, through Type I and Type II probe designs). After scanning, the raw data are stored in paired IDAT files that contain probe-level fluorescence intensities for each color channel (green and red). Illumina's DNA methylation microarrays provide genome-wide profiling of CpG methylation at scale. The Infinium MethylationEPIC v1.0 (EPICv1) array interrogates over 450,000 CpG sites, while the Infinium MethylationEPIC v2.0 (EPICv2) updates and expands probe content to increase coverage of functionally relevant loci to almost 930,000 CpG sites \cite{IlluminaEPICv1Datasheet2019}\cite{IlluminaEPICv2Datasheet2022}. 

These raw intensities in the IDAT files are read by standard software (e.g., \texttt{minfi}) to perform quality control, background correction, and normalization, and they are then transformed into methylation measures. The most common measure is the \emph{beta value}, defined as the proportion of methylated signal intensity relative to total signal, typically computed as $\beta = \frac{M}{M + U + \alpha}$, where $M$ and $U$ are the methylated and unmethylated intensities and $\alpha$ is a slight offset to stabilise estimates at low intensity. Beta values range from 0 to 1, providing a clear and intuitive representation of methylation levels.

\section{The Capper \textit{et al.} Classifier}
\label{S-Crititical}

In what follows, we critically examine the methylation-based classifier proposed by Capper et al. \cite{Capper2018}, focusing not only on the headline performance figures but, more importantly, on the methodological choices that structure the entire pipeline and the transparency with which they are justified.

\subsection{Feature Selection}

One of the foundational steps in the Capper et al. classification pipeline is dimensionality reduction of the methylation data, which originally comprised approximately 428,799 CpG probes per sample. To reduce dimensionality, the authors applied a supervised feature selection strategy based on random forest-based importance scores, ultimately selecting the top 10,000 probes as input features for model training. 

This choice is a fixed design decision, yet the article offers no theoretical or empirical justification for selecting this specific number of features. There is no indication that alternative thresholds were tested, nor is there any discussion of the trade-offs involved in retaining more or fewer probes, despite the noticeable impact such a choice has on both model complexity and biological representativeness. Given the magnitude of this step in the modeling process, this lack of transparency and methodological rigor represents a significant flaw in the classifier.

\subsection{Learning Model and Calibration Protocol}

The core classification model adopted is a random forest composed of 10,000 trees, trained with a fixed feature-subset size (\texttt{mtry}) of 100 and a downsampling strategy that limits each class to a maximum of 8 samples per tree. These choices, central to the performance and behavior of any ensemble model, are applied without discussion, benchmarking, or sensitivity analysis. There is no justification for selecting 10,000 trees (as opposed to a smaller, more computationally efficient ensemble), the fixed feature subset size, or the aggressive class-balancing strategy, which may distort class priors. The absence of hyperparameter tuning or even minimal exploration of alternatives suggests that these parameters were selected arbitrarily. This approach directly contradicts established machine learning practice.

Beyond this, the authors note that they calibrated the random forest scores but fail to explain how this was implemented. Standard calibration techniques are inherently designed for binary classification. To apply them in a multiclass context, one must either calibrate each class in a one-vs-rest scheme or calibrate only the final ``match-vs-non-match'' probability, as \cite{Capper2018} implicitly does by defining a classification score threshold ($\geq$ 0.9). However, the paper provides no details on whether calibration was applied per class, globally, or post hoc on the match confidence, leaving the entire calibration mechanism opaque. 

Calibration introduces a second learning layer, often a regression model, which may adjust for biases in the random forest's output but also risks masking systematic deficiencies in the original model. If not properly nested within the cross-validation folds, this step can lead to overfitting and overly optimistic probability estimates, particularly when applied to outputs that already benefit from leakage-prone feature selection. Without a clear description of the calibration method and its integration with validation, it is impossible to assess the trustworthiness or clinical safety of the predicted probabilities.

\subsection{Performance Reporting}

Capper \textit{et al.} report an AUC of 0.99 as a key measure of classifier performance. However, this value is derived not from the actual 91-class prediction task but from a binarised formulation in which samples are labeled as ``match'' (classification score $\geq$ 0.9) or ``non-match'' (score $<$ 0.9). The resulting ROC curve evaluates the model's ability to distinguish between high- and low-confidence predictions but does not correctly classify tumor types. This approach transforms a complex multiclass classification problem into an artificial binary task, conflating calibration confidence with classification accuracy. Such an approach may give the illusion of strong overall performance while concealing confusion among biologically and clinically distinct tumor entities.

In addition to this problematic use of AUC, the paper reports only a few global metrics—such as overall error rate (4.28\%), Brier score, sensitivity (0.989), and specificity (0.999), without providing per-class performance evaluation. In the confusion matrix, class-specific precision, recall, and F1 scores are reported, and standard multiclass evaluation strategies are not used. As a result, systematic errors are entirely hidden.

\subsection{Prospective Clinical Evaluation}

In a prospective clinical evaluation involving 1,104 cases, only 76\% of samples showed immediate concordance between the model's prediction and the original histopathological diagnosis at the 91 methylation classes. An additional 12\% of cases were initially discordant. Among these, only approximately 6\% were retrospectively reclassified to match the model's original predicted 91 classes, raising the class-level concordance to 82\%. 

However, rather than reporting this more modest performance figure, the authors shift the evaluation to a higher level of abstraction: they accept predictions that fall within the same methylation class family (MCF), a broader grouping spanning 24 tumor types, as correct. With this reframing, they report a final concordance of 88\%.

This post hoc redefinition of correctness introduces a critical inconsistency between the model's stated objective and how its performance is ultimately assessed. The clinical test is intended to validate the model's ability to identify precise tumor classes rather than broader families. By shifting the target metric during evaluation, the authors blur the distinction between accurate classification and coarse categorization, undermining interpretability and masking the model's difficulty in distinguishing closely related but clinically distinct entities.

Even more concerning is that reclassification in favor of the model implicitly suggests errors in the original histopathological labels. Nevertheless, no analogous audit is performed on the 2{,}801-sample training dataset, which the model assumes to be correct. If mislabeling occurred at similar rates in the training data, it would introduce systematic noise that would affect both the model's learning process and internal validation. Overall, the study reveals a central methodological oversight by failing to subject the training set to the same critical scrutiny applied to the validation cohort.

\section{Proposed Approach}
\label{S-Proposed} 

Following best practices in machine learning, we propose a novel multiclass classification approach for the problem we are addressing. Our approach includes dimensionality reduction, model training, performance evaluation, and model persistence, with each component tailored to address the specific challenges of DNA methylation-based classification.

\subsection{Hyperparameter Tuning}

To optimize model performance, we performed hyperparameter tuning using a 3-fold cross-validation grid search. The search was implemented via \texttt{GridSearchCV}, which exhaustively explores all specified parameter combinations using the training data.

The search space included the distortion tolerance \texttt{srp\_eps} of the SRP step and the key hyperparameters of the logistic regression classifier. Specifically, we tuned the L2-regularization strength \texttt{clf\_\_C} (with values 0.1, 1, and 10) and compared two solvers: \texttt{lbfgs} and \texttt{saga}. The penalty was fixed to \texttt{l2} throughout. The whole grid of parameters was defined as:

\begin{verbatim}
param_grid = {
    "srp__eps": [0.1, 0.2, 0.25],
    "clf__C": [0.1, 1, 10],
    "clf__solver": ["lbfgs", "saga"]
}
\end{verbatim}

We used \texttt{cv=3} to perform stratified 3-fold cross-validation for each parameter combination and selected the best configuration based on classification accuracy. The best configuration found by the grid search was \texttt{srp\_\_eps = 0.2}, \texttt{clf\_\_C = 0.1}, and \texttt{clf\_\_solver = 'saga'}. This same configuration was used in all subsequent experiments. The \texttt{n\_jobs=-1} parameter enabled full parallelization across CPU cores, and \texttt{verbose=1} provided live progress feedback during the search process.

\subsection{Dimensionality Reduction with Sparse Random Projection}

Given the extremely high dimensionality of the original data — approximately 450,000 attributes per sample, as typically found in DNA methylation profiles — we applied a dimensionality-reduction strategy as the initial step in the pipeline. The goal was to reduce computational cost, minimize the risk of overfitting, and retain the most relevant structure in the data. 

We employed \textit{Sparse Random Projection} (SRP), which maps the original data into a lower-dimensional space using a sparse random matrix. This technique is grounded in the Johnson-Lindenstrauss Lemma, which guarantees approximate preservation of pairwise distances with high probability when the projection dimension \( k \) satisfies:

\[
k \geq \frac{4 \ln(n)}{\varepsilon^2/2 - \varepsilon^3/3}
\]

In our case, with \( n = 2,801 \) samples and a tolerated distortion of \( \varepsilon = 0.2 \), the required number of projected dimensions is \( k = 1,831 \). This distortion tolerance was determined via hyperparameter tuning, balancing the trade-off between dimensionality reduction and classification accuracy. Thus, the original 450,000-dimensional data were projected onto approximately 1,831 features, while preserving the geometric relationships relevant for classification. 

\begin{verbatim}
rp = SparseRandomProjection(
    eps=0.2,
    dense_output=True
)
\end{verbatim}

\subsection{Multinomial Logistic Regression Classifier}

For the classification stage, we used multinomial logistic regression, a linear model that estimates class-membership probabilities via a softmax function applied to linear combinations of the input features.

For \(K\) classes, the model computes a discriminative score for each class \(c \in \{1,\ldots,K\}\):
\[
f_c(\mathbf{x}) = \mathbf{w}_c^\top \mathbf{x} + b_c,
\]
and the probability that a sample belongs to class \(c\) is given by the softmax function:
\[
P(y = c \mid \mathbf{x}) = 
\frac{\exp\!\left(f_c(\mathbf{x})\right)}
{\sum_{k=1}^{K}\exp\!\left(f_k(\mathbf{x})\right)}.
\]

The classifier was instantiated with:
\begin{verbatim}
clf = LogisticRegression(
    penalty="l2",
    C=0.1,
    class_weight="balanced",
    random_state=23,
    solver="saga",
    max_iter=10000,
    multi_class="multinomial"
)
\end{verbatim}

We addressed class imbalance by enabling \texttt{class\_weight="balanced"}, which adjusts the weight of each class inversely proportional to its frequency in the training set. This ensures that minority classes receive greater weight during training.

To prevent overfitting and improve generalization, we applied L2 regularization (\texttt{penalty="l2"}), which penalizes large model coefficients. We set the regularization strength to \texttt{C=0.1}, matching the grid-selected configuration reported above. We selected \texttt{solver="saga"} for its scalability in high-dimensional settings and its support for multinomial logistic regression with L2 regularization. Hyperparameters were selected via cross-validated grid search based on classification accuracy.

To ensure convergence in the high-dimensional feature space of DNA methylation data, we set \texttt{max\_iter=10000}, allowing sufficient iterations for the model to optimize. Finally, \texttt{random\_state=23} fixes the random number generator state, ensuring reproducibility across runs.
\subsection{Stratified 3-Fold Cross-Validation}

To ensure robust model evaluation while preserving class distribution, we employed stratified 3-fold cross-validation using \texttt{StratifiedKFold(n\_splits=3, shuffle=True, random\_state=23)}. This procedure divides the dataset into three mutually exclusive folds while preserving, in each fold, approximately the same class proportions observed in the full dataset.

The \texttt{shuffle=True} parameter randomizes the sample order before splitting, helping mitigate biases that might arise from the original data ordering. The \texttt{random\_state=23} parameter sets a fixed seed for the random number generator, ensuring the splits are reproducible across different runs. This stratified splitting strategy is crucial for imbalanced classification tasks, as it prevents under- or over-representation of minority classes in the training or validation sets.

\section{Experiments and Results}
\label{S-ExpResult} 

We implemented the proposed method in Google Colab to conduct the experiments and collect the results. The training and validation dataset is available in GEO (NCBI) as \href{https://www.ncbi.nlm.nih.gov/geo/query/acc.cgi?acc=GSE90496}{GSE90496}, and the clinical analysis dataset is available as \href{https://www.ncbi.nlm.nih.gov/geo/query/acc.cgi?acc=GSE109379}{GSE109379}.

The training dataset comprises 2,801 tumor samples, each represented by normalized methylation indices across approximately 450,000 CpG probes. The clinical validation set contains 1,104 samples, but each is profiled using a reduced set of roughly 42,000 probes. Each CpG probe is identified by a unique ID assigned by the manufacturer of the DNA methylation array platform. During data preparation, the training dataset was filtered to retain only probes present in the clinical validation set, ensuring feature consistency between the two datasets. As a result, only the intersection of the probe sets was used for model training and evaluation.

We report complementary multiclass metrics suited to imbalanced data, combining global measures (accuracy and weighted F1), per-class measures (balanced accuracy, macro-precision, macro-recall, and macro-F1), and imbalance-robust agreement measures (MCC and Cohen's $\kappa$).

\begin{figure}[H]
  \centering
  \includegraphics[width=0.99\columnwidth]{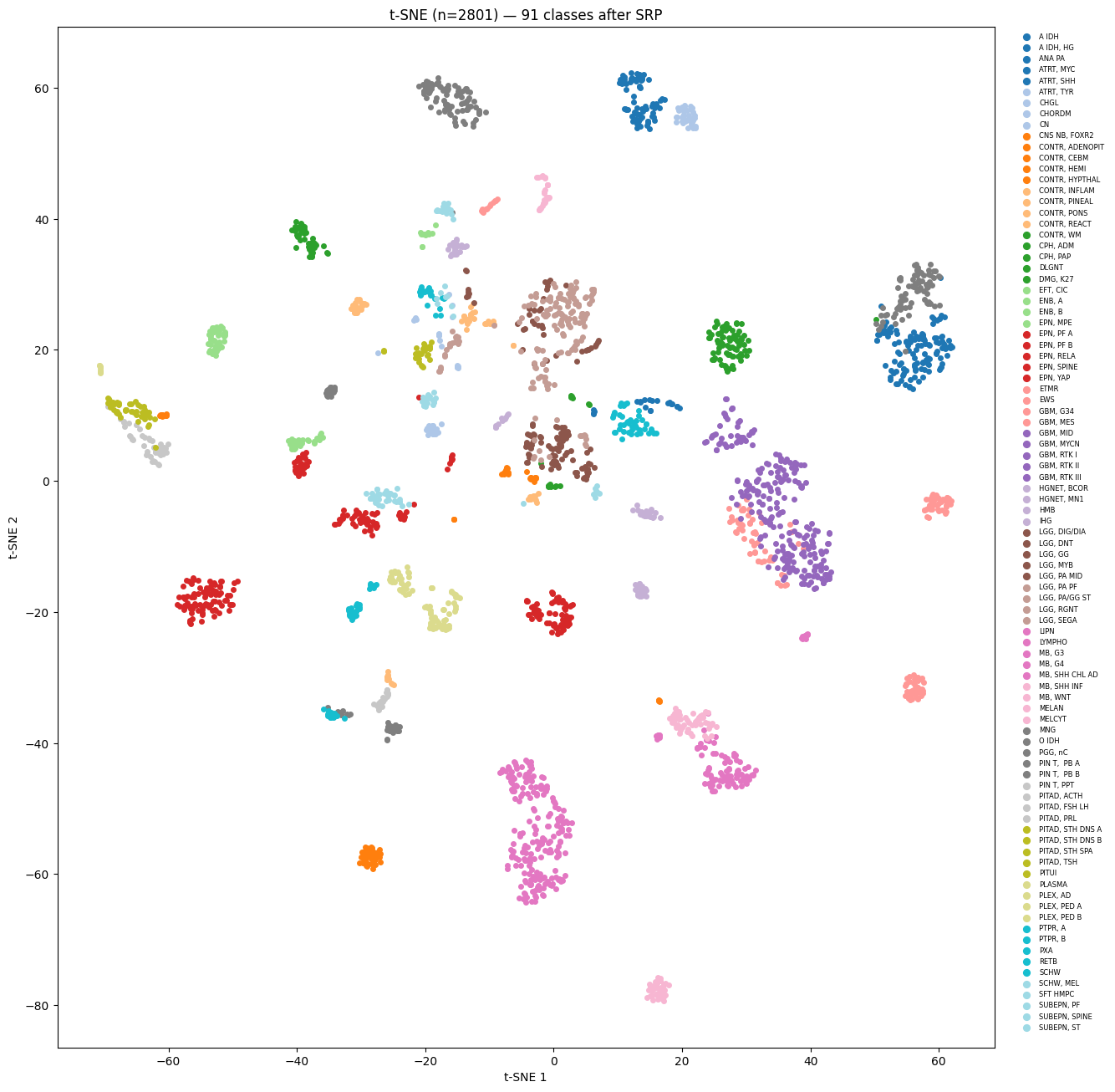}
  \caption{t-SNE embedding of the 2,801-sample reference cohort after SRP, colored by the 91 methylation classes.}
  \label{fig:tsne_91_srp}
\end{figure}

\subsection{Dimensionality Reduction}

Given the extremely high dimensionality of DNA methylation profiles, we first applied Sparse Random Projection (SRP) to map each sample to a substantially lower-dimensional space while approximately preserving pairwise distances. 

To qualitatively assess whether the SRP space preserved the cohort's intrinsic structure, we computed a two-dimensional t-SNE embedding from the SRP-projected features for visualization. Figure~\ref{fig:tsne_91_srp} shows the embedding for the whole 2,801-sample reference cohort across the 91 methylation classes. The resulting global organization is highly consistent with the class-wise arrangement previously reported for this reference dataset, prior to dimensionality reduction, as described in ~\cite{Capper2018}. Samples form compact, class-consistent islands, and biologically related entities occupy neighboring regions. This qualitative agreement supports the notion that SRP preserves the cohort geometry relevant to class separation while providing an efficient representation for downstream modeling.

\subsection{Cross Validation}

We evaluated the proposed pipeline using stratified 3-fold cross-validation on the 2,801-sample, 91-class reference cohort. The classifier achieved consistently high performance across folds (Table~\ref{tab:metrics_summary}), with an overall accuracy of 0.9658 $\pm$0.0039 (corresponding to an average error rate of 3.42\%) and a balanced accuracy of 0.9574 $\pm$0.0145, indicating strong per-class sensitivity despite the pronounced class imbalance. Macro-precision was slightly higher than macro-recall (0.9702 $\pm$0.0072 vs.\ 0.9574 $\pm$0.0145), suggesting that false positives were comparatively rare. At the same time, the remaining errors were primarily driven by a subset of harder (often smaller) classes. Chance- and imbalance-robust agreement measures (MCC and Cohen's $\kappa$) closely matched the accuracy (both $\approx$0.965), reinforcing the model's robustness in the multiclass setting.

\begin{table}[H]
\centering
\caption{Summary of quantitative performance on the reference cohort and clinical evaluation cohort.}
\label{tab:metrics_summary}
\small
\setlength{\tabcolsep}{3.5pt}
\renewcommand{\arraystretch}{0.98}
\begin{tabular}{@{}lcc@{}}
\toprule
Metric & Reference cohort (3-fold CV) & Clinical cohort \\
\midrule
Acc.             & 0.9658$\pm$0.0039 & 0.8687 \\
bAcc.            & 0.9574$\pm$0.0145 & 0.8295 \\
F1 (macro)       & 0.9593$\pm$0.0124 & 0.7608 \\
F1 (weighted)    & 0.9645$\pm$0.0042 & 0.8707 \\
Prec. (macro)    & 0.9702$\pm$0.0072 & 0.7813 \\
Rec. (macro)     & 0.9574$\pm$0.0145 & 0.7632 \\
MCC              & 0.9652$\pm$0.0039 & 0.8619 \\
Cohen's $\kappa$ & 0.9651$\pm$0.0039 & 0.8617 \\
\bottomrule
\end{tabular}
\end{table}

\begin{figure}[H]
  \centering
  \includegraphics[width=\columnwidth]{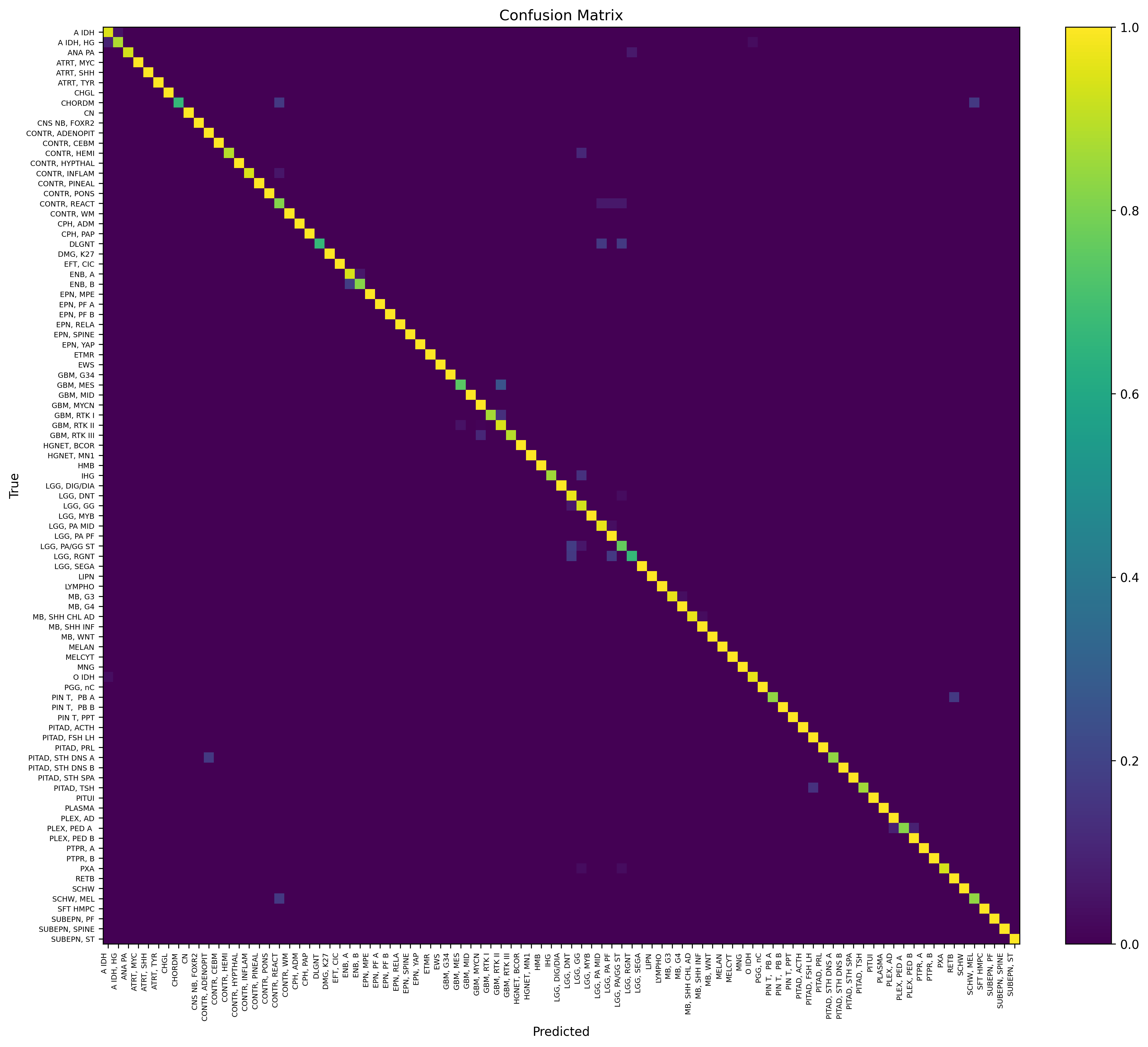}
  \caption{Confusion matrix for the 91-class (3-fold CV).}
  \label{fig:cm_oof}
\end{figure}

Figure~\ref{fig:cm_oof} summarizes the 91-class cross-validated confusion matrix (row-normalized). As expected in an extensive taxonomy of closely related CNS tumor entities, most off-diagonal mass was concentrated within biologically and histologically neighboring classes rather than scattered across distant entities. The predominant misclassifications occurred within coherent subfamilies, including glioblastoma subtypes (e.g., RTK I/II/MES/MYCN), IDH-mutant diffuse glioma-related entities (A IDH, O IDH, and A IDH, HG), and low-grade glioma categories that are known to share overlapping methylation features. Notably, the overall structure indicates that the classifier rarely confuses samples from completely unrelated diagnostic groups. When errors occur, they tend to cluster within clinically plausible neighborhoods in methylation space.

Capper \textit{et al.} reported cross-validated error rates of 4.89\% (raw) and 4.28\% (calibrated) for the same 2,801-sample / 91-class reference cohort using their classifier. In our stratified 3-fold cross-validation, the proposed approach yields an average error rate of 3.42\%, which is comparable to—and slightly lower than — those values. While our study uses a different modeling choice, the qualitative error structure observed here similarly concentrates within biologically adjacent entities, supporting the view that remaining errors in large CNS methylation taxonomies often reflect intrinsic proximity between related classes rather than arbitrary mislabelling across distant diagnostic groups.

\subsection{Clinical Evaluation}

We next evaluated the proposed model on the independent clinical evaluation cohort originally used by Capper \textit{et al.} as a prospective diagnostic series. This cohort is inherently more challenging than the reference cohort cross-validation setting, as it is enriched for rare and diagnostically complex cases and includes real-world variability in sample quality and tumor cell content. In the original study, 51 out of 1,155 cases (4\%) were not suitable for methylation profiling. Among the 1,104 profiled samples, only a subset achieved a high-confidence match based on a calibrated score threshold \cite{Capper2018}.

On this external cohort, our method achieved an overall accuracy of 0.8687 and balanced accuracy of 0.8295 (Table~\ref{tab:metrics_summary}). The larger gap between weighted and macro-averaged metrics (F1$_\mathrm{weighted}$=0.8707 vs.\ F1$_\mathrm{macro}$=0.7608) indicates that a substantial fraction of the residual error is driven by the long tail of low-support classes, which is expected in a 91-class clinical taxonomy under distribution shift.

Importantly, the headline figure reported by Capper \textit{et al.} for the prospective cohort should not be interpreted as a 91-way test-set accuracy. They report that 88\% of profiled samples (977/1{,}104) ``match'' a methylation class using a thresholded decision rule (common calibrated score cutoff $\ge 0.9$), including a family-aware relaxation for subclasses within methylation class families (subclass score $\ge 0.5$ accepted if the summed family score is $\ge 0.9$) \cite{Capper2018}. 

\begin{figure}[H]
\centering
\includegraphics[width=0.99\columnwidth]{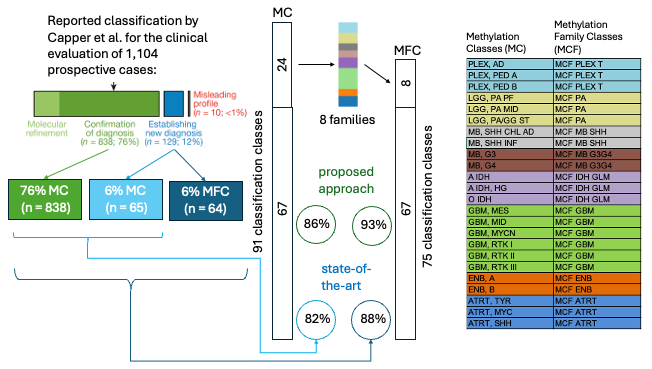}
\caption{Clinical evaluation of the proposed approach vs. Capper \textit{et al.} (state-of-the-art)}
\label{fig:clinical}
\end{figure}

Under a strict class-level concordance criterion \emph{prior} to any re-evaluation of discordant cases, the agreement between histopathology and methylation profiling is 76\% (838/1{,}104). As summarized in Fig.~\ref{fig:clinical}, even if one credits only the subset of additional cases that can be considered resolved at the \emph{class level} without resorting to the family-level relaxation (approximately +6 percentage points in our reconstruction), the class-level concordance would rise only to about 82\%. The higher 88\% number is achieved under a broadened evaluation endpoint, which changes the granularity and definition of success relative to the original 91-class classification task.

In Fig.~\ref{fig:clinical}, we also contrast these reporting conventions with our fixed 91-class evaluation on the same clinical cohort, where every case receives a class assignment and we compute standard multiclass metrics (Table~\ref{tab:metrics_summary}). Under this stricter and fully multiclass criterion, our method achieves an accuracy of 86.9\%. This is contrasted with an effective upper bound of approximately 82\% for the original Capper \textit{et al.} class-level concordance once the family-level relaxation is removed (i.e., retaining only cases that can reasonably be counted as correctly resolved at the \emph{class} granularity without collapsing errors into broader groupings).

For completeness, we also report the results obtained when adopting the same relaxed, family-aware interpretation used in the Capper clinical analysis. Specifically, when we evaluate concordance at the level of methylation class families (MCFs), treating within-family confusions as concordant, our predictions in the clinical cohort reach approximately 93\% concordance. This highlights that a substantial portion of the remaining discrepancies at the 91-class level are concentrated within biologically related groups, and that the choice of evaluation granularity (class vs.\ family) can materially change the headline performance figure.

Overall, on this clinically critical external validation cohort, the proposed approach outperformed the state of the art under both evaluation regimes. Under a strict 91-class criterion, our method achieves higher class-level performance (86.9\% vs. 82\%), and under the relaxed family-level interpretation analogous to Capper \textit{et al.}, our reconstructed MCF-level concordance is also higher (93\% vs. 88\%).

\section{Conclusion}
\label{S-conclusion} 

This article proposed a simple, fully reproducible, and computationally efficient pipeline for DNA methylation-based brain tumor classification, designed to be transparent and practical for clinical translation. The method combines Sparse Random Projection for dimensionality reduction with a multinomial logistic regression classifier trained on the 91-class reference cohort. By relying on a distance-preserving projection and a linear decision model, the approach aims to retain discriminative structure while avoiding the complexity and opacity of more elaborate architectures. Despite its simplicity, the proposed approach achieved strong and stable performance on the 91-class reference cohort under stratified 3-fold cross-validation (Acc.\ $\approx$96.6\%), indicating that a linear decision model can be highly competitive when paired with an appropriate distance-preserving projection.

On the independent clinical evaluation cohort originally used by Capper \textit{et al.}, our method achieved higher performance under a strict 91-class evaluation criterion (Acc.\ $\approx$ 86.9\%). Significantly, when adopting the same relaxed, family-level interpretation used in the prior state-of-the-art analysis, our reconstructed MCF-level concordance also increased (to $\approx$93\%), exceeding the headline family-level figure reported in that framework (88\%). Together, these results support the proposition that the proposed approach is competitive under a like-for-like class-level evaluation and remains strong even under a broader family-level criterion.

We make an online implementation of our pipeline available for research use only at
\url{https://genotipo.com.br/methylation/cnst_classifierv3.php}\\
(\textit{Central Nervous System Tumour Methylation Classifier}). The tool implements the SRP + Logistic Regression pipeline described in this article, trained on the 2,801-sample reference cohort from Capper \textit{et al.} (2018), and outputs predictions across the same 91 CNS tumour methylation classes. Users can upload Illumina IDAT files from EPICV2 beadchip to obtain methylation-class predictions.

Future work will focus on two complementary directions. First, we will further investigate the dimensionality-reduction stage to enhance interpretability and robustness by examining alternatives and hybrid strategies that better preserve probe-level information while retaining the computational advantages needed for high-dimensional methylation profiles. Second, we will broaden external validation by testing the approach on larger multicenter cohorts covering additional CNS tumors, enabling a more realistic assessment of reliability, calibration, and clinical utility across heterogeneous laboratory and patient contexts.

\section*{Acknowledgments}

The authors acknowledge financial support from the Conselho Nacional de Desenvolvimento Científico e Tecnológico (Chamada CNPq/MCTI/FNDCT 22/2024, Processo 444940/2024-3) and the Kunumi Institute, and thank both institutions for their commitment to advancing scientific research.

{\footnotesize
\bibliographystyle{splncs04}
\bibliography{biblio}
}

\end{document}